\documentclass[runningheads]{llncs}

\usepackage{eccv}

\usepackage{eccvabbrv}

\usepackage{graphicx}
\usepackage{booktabs}

\usepackage[accsupp]{axessibility}  

\usepackage{hyperref}

\usepackage{orcidlink}

\title{Dual-Output Multi-Exposure HDR Reconstruction via SDR Fusion and Gain Map Inverse Tone Mapping} 

\titlerunning{DOME-HDR}

\author{
Jinho Kim\,\orcidlink{0009-0000-2055-8683}
\and
Jinwoo Kim\,\orcidlink{0009-0001-3250-1788}
\and
Seon Joo Kim\,\orcidlink{0000-0001-8512-216X}
}

\authorrunning{J. Kim et al.}

\institute{Yonsei University, Republic of Korea}

\begin{document}
\maketitle

\begin{abstract}
We propose DOME-HDR, a dual-output multi-exposure HDR reconstruction framework that jointly produces a perceptually balanced SDR image and a consistent HDR image via gain map inverse tone mapping. Given three bracketed LDR inputs, DOME-HDR first synthesizes a base SDR using a LoRA-adapted latent diffusion model. A dual cross-attention fusion module injects complementary structural and color cues from the under- and over-exposed images while anchoring on the mid exposure for stability. The synthesized SDR then guides HPGM, our HDR Prior-guided Gain Map network, to predict a spatially varying gain map for reliable dynamic-range expansion. We evaluate on Kalantari, Tel, and Challenge123 using both full-reference and no-reference metrics, where DOME-HDR achieves state-of-the-art HDR reconstruction quality; ablations further confirm the effectiveness of dual cross-attention and SDR-guided gain map estimation.
\end{abstract}    
\section{Introduction}
\label{sec:intro}

\begin{figure}[t]
    \centering
    \includegraphics[width=\linewidth]{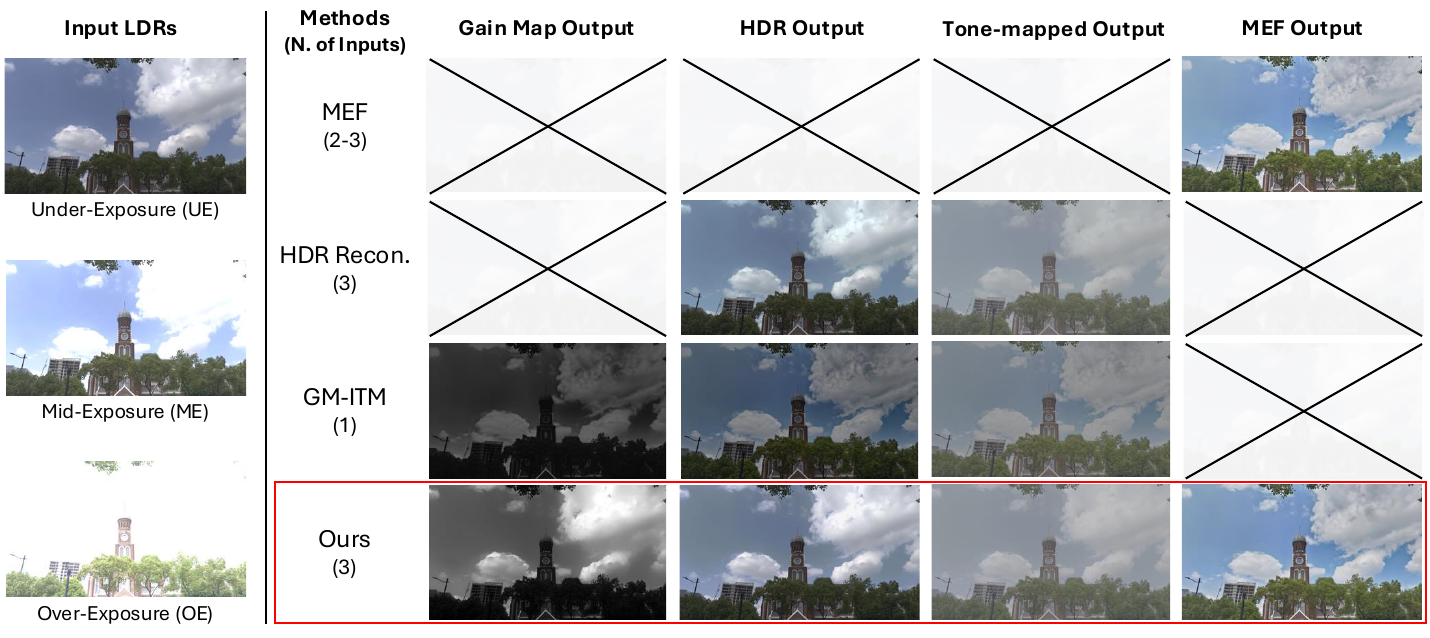}
    \caption{
Comparison of dynamic range expansion strategies. 
Existing multi-exposure fusion (MEF)\cite{chen2025ultrafusion}, multi-exposure HDR reconstruction\cite{li2025afunet}, and gain map-based method\cite{liao2025learning} operate in disjoint domains and produce partial outputs. 
In contrast, our framework jointly predicts MEF, gain map, tone-mapped SDR, and HDR using a single framework.
For visualization, HDR results are tone-mapped in the PQ domain.
}
    \label{fig:overview}
\end{figure}

The dynamic range of an image determines how much visual information can be preserved from real-world scenes. 
Natural environments often exhibit extreme luminance variations, covering several orders of magnitude between dark shadows and bright highlights. 
However, conventional Standard Dynamic Range (SDR) imaging systems are inherently constrained by limited capture and display capabilities, preventing accurate representation of wide luminance distributions.

To overcome this limitation, recent research efforts have focused on dynamic range expansion through multi-exposure fusion (MEF) and HDR reconstruction. 
MEF\cite{mertens2007exposure,li2014selective,li2017detail,li2020fast,ma2017robust,jiang2023meflut,liu2023emef,wu2024hsds,zhu2024tcmoa,chen2025ultrafusion} combines multiple Low Dynamic Range (LDR) images captured at different exposure levels into a single perceptually enhanced result. 
Recent diffusion-based MEF approach\cite{chen2025ultrafusion} has demonstrated strong performance in generating visually pleasing images with improved contrast and detail preservation. 
However, the fused output remains bounded within SDR representation and does not explicitly recover scene radiance beyond SDR constraints.

HDR reconstruction\cite{chen2021hdrunet, wang2022kunet, kim2024dcdrunet,kalantari2017deep, wu2018deep, yan2019attention, yan2020nonlocal, liu2021adnet, ye2021psfnet, liu2022cavit, chen2023hyhdr, yan2023diffhdr, tel2023alignment, hu2024frequency, li2025afunet} instead aims to recover high dynamic range radiance representations. 
Single-image HDR reconstruction\cite{chen2021hdrunet, wang2022kunet, kim2024dcdrunet} predicts HDR content from a single SDR input, typically relying on generative priors to compensate for missing information. However, such approaches rely on a single observation, lacking complementary exposure cues and making the estimation under-constrained. In contrast, multi-exposure HDR reconstruction\cite{kalantari2017deep, wu2018deep, yan2019attention, yan2020nonlocal, liu2021adnet, ye2021psfnet, liu2022cavit, chen2023hyhdr, yan2023diffhdr, tel2023alignment, hu2024frequency, li2025afunet} leverages complementary information from differently exposed inputs and reconstructs HDR images by aligning and fusing multiple LDR observations. By integrating complementary exposure information, multi-exposure methods reconstruct radiance using details preserved across different exposure levels.

While these methods achieve strong performance, they exhibit several limitations. First, they often require HDR-domain modeling, since many modern generative backbones such as VAE or diffusion-based models are designed and pretrained in the SDR domain. Extending these frameworks to HDR reconstruction therefore demands additional HDR-specific adaptation strategies. Second, HDR reconstruction is commonly performed in the linear domain and requires subsequent tone mapping for visualization. This makes the final perceptual quality dependent on the chosen tone mapping operator and target display characteristics, causing perceptual variations or degradation across display devices.

As an alternative formulation, Gain Map-based Inverse Tone Mapping (GM-ITM)\cite{liao2025learning,canham2025gainmlp} models the transformation between SDR and HDR as a pixel-wise residual. 
Formally, a gain map\cite{apple2021edr,adobe2024gainmap,google2024ultrahdr} can be interpreted as a multiplicative residual defined as the luminance ratio between corresponding HDR and SDR pixels. 
Instead of directly synthesizing HDR content, these methods predict a gain map representing luminance scaling between SDR and HDR images. 
By operating primarily in the SDR domain, gain map approaches avoid HDR-domain generative modeling and enable display-adaptive HDR rendering without fixed tone mapping complexity. 
Recent studies\cite{liao2025learning,canham2025gainmlp,hu2025gmodiff} have explored learning-based gain map modeling, ranging from lightweight MLP encoders to diffusion-based gain map refinement methods. Instead of directly generating HDR images, these approaches improve the fidelity and efficiency of residual reconstruction within the gain map framework.

However, a key limitation of existing gain map methods is that they implicitly assume input SDR images sufficiently preserve scene structure and highlight details. 
When the input SDR image is over-exposed or contains saturated regions, radiometric information is fundamentally lost. 
Since a gain map only scales existing signals, it cannot reconstruct missing content in clipped areas, and the resulting HDR quality is inherently limited by the quality of the base SDR input. 
One simple solution to this limitation is to combine a base SDR image generator with a gain map estimator, where the generator first enhances or reconstructs a visually improved SDR image and the gain map is then predicted on top of it. 
However, we observe that naively cascading these two components can result in suboptimal performance. 
Without joint optimization, errors introduced in the SDR generation stage propagate to the gain map prediction, limiting the overall HDR reconstruction quality.

To this end, we propose Dual-Output Multi-Exposure HDR reconstruction (DOME-HDR), a unified framework that converts a three-exposure bracket into (i) a perceptually balanced, display-ready SDR image and (ii) a corresponding HDR image within a single pipeline. DOME-HDR decomposes the problem into two connected components that are optimized jointly. 
First, we synthesize a high-quality base SDR image using a LoRA\cite{hu2022lora}-adapted latent diffusion model built upon UltraFusion\cite{chen2025ultrafusion}. 
Taking the mid-exposure as the anchor, we introduce a dual cross-attention design that injects complementary structural and color cues from the under- and over-exposed inputs through separate attention branches.
Second, the synthesized SDR image serves as a reliable reference for HPGM, our HDR Prior-guided Gain Map network, which predicts a spatially varying gain map. 
This dual-output design provides stable SDR visualization without committing to a particular tone mapping operator, while simultaneously delivering an HDR representation suitable for HDR-capable displays. 
As a result, DOME-HDR bridges these previously disjoint HDR imaging paradigms and predicts multiple outputs (SDR, gain map, and HDR) within a single framework (\cref{fig:overview}).

We validate DOME-HDR on three standard multi-exposure HDR benchmarks with both full-reference\cite{azimi2021pu21,zhang2018lpips, ding2020dists,wang2004ssim} and no-reference\cite{wang2023clipiqa, mittal2012niqe, mittal2012brisque, yang2022maniqa} evaluation. Across datasets, DOME-HDR achieves state-of-the-art HDR reconstruction on perceptual-uniform fidelity and perceptual similarity metrics, while also producing competitive SDR fusion quality compared to dedicated MEF methods. Extensive ablations further verify the effectiveness of our dual cross-attention design and training strategy.
\section{Related Works}

\subsection{Multi-Exposure Fusion}
Multi-exposure fusion (MEF)\cite{mertens2007exposure,li2014selective,li2017detail,li2020fast,ma2017robust,jiang2023meflut,liu2023emef,wu2024hsds,zhu2024tcmoa,chen2025ultrafusion} aims to integrate multiple Low Dynamic Range (LDR) images captured at different exposure levels into a single perceptually well-exposed result. 
Early MEF methods\cite{mertens2007exposure,zhang2011gradient} relied on hand-crafted weight maps in the spatial or transform domain to fuse images via weighted averaging. 
With the advent of deep learning, CNN-based models\cite{prabhakar2017deepfuse,xu2020mefgan,xu2020u2fusion,liang2022fusion} replaced heuristic rules with data-driven fusion strategies, followed by more recent deep architectures that improved global context modeling and detail preservation. 
Recently, diffusion-based framework such as UltraFusion~\cite{chen2025ultrafusion} has further enhanced perceptual realism by leveraging generative priors. It formulates exposure fusion as a guided inpainting problem, using the mid-exposure image as the reference while incorporating complementary information from other exposures to recover saturated regions and produce visually plausible fusion results.

Despite these improvements, MEF fundamentally operates in the LDR domain and prioritizes perceptual exposure balance rather than radiance reconstruction. 
In contrast to conventional MEF, we consider fusion as an intermediate representation rather than the final objective. In our framework, MEF is used to construct a high-quality base SDR image, which subsequently guides gain map prediction for HDR reconstruction with expanded luminance range.

\subsection{HDR Reconstruction}
High Dynamic Range (HDR) Reconstruction\cite{chen2021hdrunet, wang2022kunet, kim2024dcdrunet, kalantari2017deep, wu2018deep, yan2019attention, yan2020nonlocal, liu2021adnet, ye2021psfnet, liu2022cavit, chen2023hyhdr, yan2023diffhdr, tel2023alignment, hu2024frequency, li2025afunet} aims to recover a radiometrically accurate HDR image from one or multiple Low Dynamic Range (LDR) inputs. Unlike multi-exposure fusion, HDR reconstruction explicitly targets the recovery of scene radiance beyond the limited dynamic range of individual LDR images, producing a physically meaningful representation.

HDR reconstruction methods are broadly categorized into single-image and multi-exposure approaches. 
Single-image methods\cite{chen2021hdrunet, wang2022kunet, kim2024dcdrunet} infer HDR content from a single LDR input using learned priors, but remain fundamentally limited in recovering severely clipped highlights due to irreversible information loss. 
In contrast, multi-exposure methods\cite{kalantari2017deep, wu2018deep, yan2019attention, yan2020nonlocal, liu2021adnet, ye2021psfnet, liu2022cavit, chen2023hyhdr, yan2023diffhdr, tel2023alignment, hu2024frequency, li2025afunet} exploit complementary information from differently exposed images.
Recent deep learning–based models further integrate alignment and reconstruction within unified frameworks, improving robustness to motion and saturation. 
For instance, SAFNet~\cite{kong2024safnet} introduces selective alignment to adaptively fuse exposure features, while SCTNet~\cite{tel2023sctnet} employs transformer-based cross-attention to enforce semantic consistency during alignment. 
More recently, AFUNet~\cite{li2025afunet} formulates HDR reconstruction within a deep unfolding framework, jointly modeling alignment and fusion through cross-iterative optimization.

In our work, we revisit HDR reconstruction from a different perspective. Rather than directly predicting linear-domain HDR radiance as the final output, we aim to reformulate the reconstruction process through a gain map–based representation built upon a high-quality base SDR image. This design allows us to retain the structural advantages of multi-exposure images while avoiding direct HDR-domain synthesis as the primary objective.

\subsection{Gain Map-based Inverse Tone Mapping}
Gain Map-based Inverse Tone Mapping (GM-ITM) has emerged as a practical framework for backward-compatible HDR delivery. Instead of a full HDR image, the method encodes an SDR base image together with a spatially varying residual transformation, referred to as a gain map. Given an SDR image S(x,y) and its HDR counterpart H(x,y), the gain map is defined as a pixel-wise ratio:

\begin{equation}
f(x,y) = \frac{H(x,y) + \epsilon}{S(x,y) + \epsilon}
\label{eq:gain_def}
\end{equation}
where $\epsilon$ ensures numerical stability. HDR reconstruction is then performed as
\begin{equation}
\hat{H}(x,y) = (\hat{S}(x,y) + \epsilon) \odot \hat{f}(x,y) - \epsilon,
\label{eq:hdr_recon}
\end{equation}
where $\odot$ denotes element-wise multiplication.

Under this formulation, GM-ITM focuses on predicting the spatially varying multiplicative residual $f(x,y)$ rather than directly regressing the HDR image itself. Since HDR reconstruction in Eq.~\eqref{eq:hdr_recon} is achieved through luminance expansion of the SDR signal, the problem is cast in the rendered image domain, modeling how the SDR representation should be scaled to HDR appearance.

Learning-based GM-ITM methods~\cite{liao2025learning,canham2025gainmlp,hu2025gmodiff} directly predict the gain map as the reconstruction target. 
A recent method adopts a dual-branch design~\cite{liao2025learning} to capture both local contrast variations and global luminance scaling, while coordinate-based networks~\cite{canham2025gainmlp} model the gain map as a continuous function conditioned on spatial location and SDR content.
Diffusion-based refinements have also been introduced to improve perceptual quality of the HDR results~\cite{hu2025gmodiff}.

Despite these advantages, the formulation in Eq.~\eqref{eq:hdr_recon} reveals an intrinsic limitation. The gain map only amplifies existing SDR signals. Consequently, information lost due to saturation or clipping in the SDR image cannot be fully recovered, making the reconstruction quality fundamentally bounded by the fidelity of the SDR base.

Motivated by this limitation, we investigate an alternative formulation that does not rely solely on single-SDR residual estimation. By incorporating additional exposure information prior to gain prediction, our method improves HDR reconstruction while maintaining compatibility with the gain map framework.

\begin{figure}[t]
    \centering
     \includegraphics[width=\linewidth]{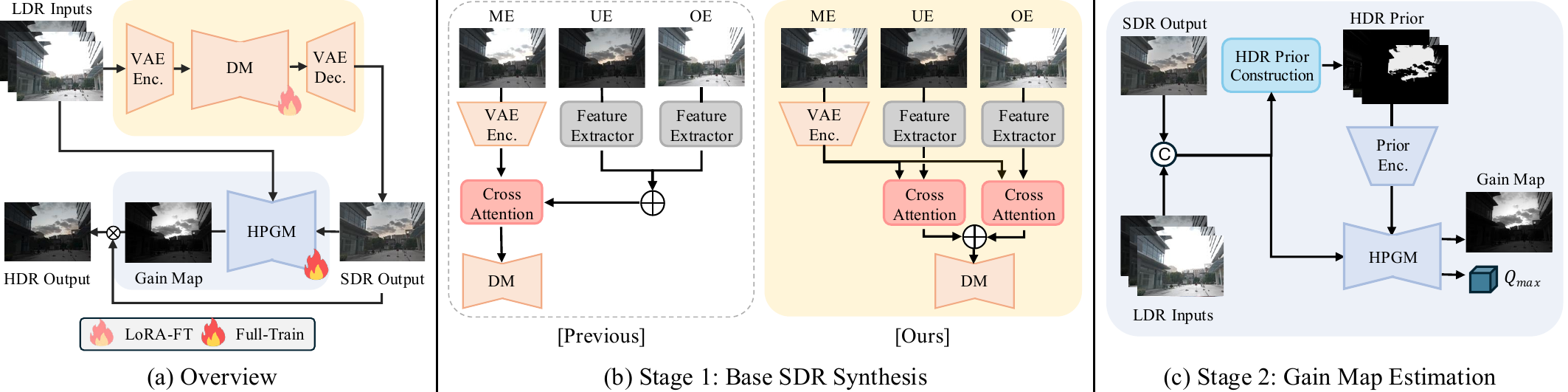}
    \caption{
Overview of the proposed framework. Given three exposure images, our method first synthesizes a visually balanced base SDR image through a diffusion-based multi-exposure fusion stage. The fused SDR image is then used to estimate a spatially varying gain map with the proposed HDR prior–guided gain map network.
}
    \label{fig:pipeline}
\end{figure}
\section{Methodology}

\subsection{Overview}

\cref{fig:pipeline}(a) demonstrates the overall pipeline of the DOME-HDR framework.
Given three LDR images captured at different exposure levels, under-exposed $I_{ue} \in \mathbb{R}^{H\times W\times3}$, mid-exposed $I_{me} \in \mathbb{R}^{H\times W\times3}$, and over-exposed $I_{oe} \in \mathbb{R}^{H\times W\times3}$, our goal is to generate both a visually pleasing SDR image $\hat{I}_{sdr} \in \mathbb{R}^{H\times W\times3}$ and a corresponding HDR image $\hat{I}_{hdr} \in \mathbb{R}^{H\times W\times3}$ while preserving perceptual consistency across exposure regions.

To achieve this goal, our framework decomposes the problem into two complementary stages by disentangling perceptual exposure fusion from HDR expansion. This design enables the proposed framework to produce a visually pleasing SDR image while simultaneously enabling HDR reconstruction.

First, to synthesize a perceptually balanced base SDR image $\hat{I}_{sdr}$ from the given inputs, we employ a diffusion-based multi-exposure fusion (MEF) module\cite{chen2025ultrafusion}. The diffusion model takes $I_{me}$ as the primary structural reference while incorporating complementary structural and color cues from $I_{ue}$ and $I_{oe}$.
Second, a gain map network predicts a spatially varying gain map $\hat{g} \in \mathbb{R}^{H\times W\times3}$  using the previously fused $\hat{I}_{sdr}$.
The predicted gain map expands the dynamic range of the fused SDR image and reconstructs the final HDR image $\hat{I}_{hdr}$.
Then, the HDR image is reconstructed using the predicted gain map as
\begin{equation}
\hat{I}_{hdr}(x,y) =
\left(\hat{I}_{sdr}(x,y) + \epsilon\right)^{\hat{g}(x,y)} - \epsilon ,
\end{equation}
where $\epsilon$ is a small constant for numerical stability.

\subsection{Stage 1: Base SDR Synthesis}

Our SDR synthesis module builds upon UltraFusion\cite{chen2025ultrafusion}, which models MEF as a guided inpainting problem. Rather than directly merging exposure brackets, UltraFusion uses the over-exposed image as a reference and inpaints its clipped highlight regions using structural and color cues decomposed from the under-exposed image, which are injected via a cross-attention mechanism.

Building upon this design, we extend the UltraFusion framework to three input exposures by using the mid-exposure image $I_{me}$ as the primary structural reference. To incorporate complementary information from the under- and over-exposed images, we extract exposure-specific structural and color cues from $I_{ue}$ and $I_{oe}$, which serve as guidance features for the diffusion model.

These guidance features are integrated through the cross-attention mechanism, as illustrated in Fig.~\ref{fig:pipeline}(b). 
In UltraFusion, extending the framework to more than two exposures is achieved by extracting guidance features from different exposures and combining them through normalized summation before injecting them into the cross-attention module. However, this aggregation may lead to information loss, as complementary cues from different exposure levels are averaged before attention and cannot be selectively utilized.

Accordingly, we employ two independent cross-attention branches that inject exposure-specific guidance features into the mid-exposure latent representation. 
For each exposure branch, structural and color features extracted from $I_{ue}$ and $I_{oe}$ are fused by concatenation followed by a $1{\times}1$ convolution to produce guidance features. 
These features serve as attention keys and values, while the latent feature derived from $I_{me}$ acts as the query in a cross-attention operation:

\begin{equation}
X_{out} = X_{me} + \text{Conv}_{1\times1}\!\left(
\text{softmax}\!\left(\frac{QK^\top}{\tau}\right)V
\right),
\end{equation}

where $X_{me}, X_{out} \in \mathbb{R}^{\frac{H}{8} \times \frac{W}{8} \times 4}$ denote the latent representation of the mid-exposure image and the updated latent feature, respectively.
The query $Q \in \mathbb{R}^{N \times C}$ is derived from the mid-exposure latent feature, where $N $ denotes the number of spatial tokens at level $l$. The keys and values $K,V \in \mathbb{R}^{N \times C}$ are computed from the fused exposure guidance features.
The operation is applied independently to the UE and OE branches, allowing structural and color cues from each exposure to be selectively transferred to the latent representation. 
During training, the ControlNet is adapted using LoRA\cite{hu2022lora} applied to the attention query and value projections, while the U-Net and VAE remain frozen.

Given these conditions, the latent diffusion model performs iterative denoising conditioned on the mid-exposure latent and the exposure-specific guidance features. The final denoised latent is decoded by the frozen VAE decoder to produce the base SDR image $\hat{I}_{sdr}$, which serves as the input for the subsequent gain map estimation network.

\subsection{Stage 2: Gain Map Estimation and Inverse Tone Mapping}
Given the synthesized base SDR image $\hat{I}_{sdr}$ and the three input LDR exposures, the second stage estimates a spatially varying per-pixel gain map $\hat{g}$ to reconstruct the final HDR image. 
We introduce the HDR Prior-guided Gain Map network (HPGM), a U-Net-based model that leverages an HDR prior to stabilize learning and predicts a residual over the prior as shown in Fig.~\ref{fig:pipeline}(c).

\subsubsection{HDR Prior Construction.}

To provide a reliable initialization for gain map estimation, we construct an HDR prior 
$G_\text{prior} \in \mathbb{R}^{H \times W \times 3}$ from the input exposure stack.
Since the input exposures capture complementary radiometric information across different exposure levels, they provide useful cues for approximating the underlying scene radiance. 
To enable consistent radiometric comparison across exposures, each image is first linearized by gamma decoding with $\gamma = 2.2$ and exposure normalization:
\begin{equation}
X_k = \frac{I_k^{\gamma}}{2^{ev_k}}, 
\quad k \in \{ue, me, oe\},
\end{equation}
where $X_k \in \mathbb{R}^{H \times W \times 3}$ denotes the linearized radiance image and $ev_k$ denotes the exposure value.

To capture cross-exposure relationships, we compute exposure ratios 
$R_{ue} = X_{ue}/X_{me}$ and $R_{oe} = X_{oe}/X_{me}$. 
Soft saturation masks are estimated from the luminance of the mid-exposure image as
\begin{equation}
S_{hi} = \sigma(\alpha \cdot (l_{me} - t_{hi})), \quad
S_{lo} = \sigma(\alpha \cdot (t_{lo} - l_{me})),
\end{equation}
where $S_{hi}, S_{lo} \in \mathbb{R}^{H \times W \times 1}$ denote the highlight and shadow masks, $l_{me}$ denotes the luminance of the mid-exposure image $I_{me}$, $t_{hi}$ and $t_{lo}$ denote predefined highlight and shadow thresholds, set to $0.95$ and $0.05$, respectively, $\sigma(\cdot)$ denotes the sigmoid function, and $\alpha$ controls the sharpness of the mask transition and is set to $50$.

Using these masks, a saturation-aware linear HDR estimate is then computed as
\begin{equation}
\hat{I}_{\text{hdr}}^{prior} =
S_{hi} \cdot X_{ue}
+ (1 - S_{hi} - S_{lo}) \cdot X_{me}
+ S_{lo} \cdot X_{oe}.
\end{equation}
In this formulation, the under-exposed image contributes in highlight regions, the over-exposed image in shadow regions, and the mid-exposure image in well-exposed regions.

The HDR prior is finally derived from the relationship between this blended HDR estimate and the base SDR image $\hat{I}_{sdr}$:

\begin{equation}
G_\text{prior} =
\frac{\log(\tilde{I}_{hdr}^{prior} + \epsilon)}
{\log(\hat{I}_{sdr} + \epsilon)},
\end{equation}

where $\tilde{I}_{hdr}^{prior}$ denotes the normalized HDR estimate. This HDR prior provides an initial estimate of the gain map and is used to guide the subsequent gain map estimation network.

\subsubsection{HDR Prior-guided Gain Map network (HPGM).}

As illustrated in Fig.~\ref{fig:pipeline}(c), HPGM is implemented as a U-Net that takes an 11-channel input formed by concatenating the base SDR image $\hat{I}_{sdr}$, the cross-exposure ratios $R_{ue}$ and $R_{oe}$, and two saturation masks $S_{hi}$ and $S_{lo}$.

The encoder extracts hierarchical features from the input representation. 
At the bottleneck, an HDR prior feature $f_\text{prior}$ is extracted by a lightweight convolutional encoder from the concatenation of $G_\text{prior}$, $S_{hi}$, and $S_{lo}$. 
This prior feature is injected into the bottleneck representation through a residual connection to guide gain map estimation.

The decoder then progressively recovers the spatial resolution through three edge-guided upsampling blocks, where Sobel-based edge maps are used as spatial gates to sharpen feature boundaries during upsampling. 
The final decoder feature is used to predict both a gain map and a global HDR scale through two prediction heads. 
The gain map head predicts a residual $\Delta g$ over the HDR prior, yielding $\hat{g} = G_\text{prior} + \Delta g$, while the scale head predicts the global HDR scale $\hat{Q}_\text{max}$ via global average pooling followed by a two-layer MLP. 
The final HDR image is reconstructed as
\begin{equation}
\hat{I}_{hdr}(x,y) =
(\hat{I}_{sdr}(x,y) + \epsilon)^{\hat{g}(x,y)} \cdot \hat{Q}_\text{max} - \epsilon .
\end{equation}

\subsection{Training Loss}

The proposed framework is trained by jointly supervising image-level reconstruction and gain-related constraints. 
Let $\hat{I}_{hdr}$ denote the reconstructed HDR image and $I_{hdr}$ the ground-truth HDR target.

We first define the image-level supervision as
\begin{equation}
\mathcal{L}_{\text{img}} =
\left\| \mathcal{T}_{\mu}(I_{hdr}) - \mathcal{T}_{\mu}(\hat{I}_{hdr}) \right\|_{1}
+ \lambda_{\text{perc}}
\left\| \phi(I_{hdr}) - \phi(\hat{I}_{hdr}) \right\|_{1},
\end{equation}
where $\mathcal{T}_{\mu}(x) = \frac{\log(1 + \mu x)}{\log(1 + \mu)}$ with $\mu = 5000$, and $\phi(\cdot)$ denotes VGG19 features. We set $\lambda_{\text{perc}} = 10^{-2}$.

The gain-level supervision is defined as
\begin{equation}
\mathcal{L}_{\text{gain}} =
\sqrt{(\hat{g} - g)^2 + \varepsilon^2}
+ w_{\text{qmax}}
\left\| \hat{Q}_{\max} - Q_{\max} \right\|_{1},
\end{equation}
where $g$ denotes the ground-truth gain map and $\varepsilon = 10^{-3}$. We set $w_{\text{qmax}} = 0.3$.

The final objective is
\begin{equation}
\mathcal{L}_{\text{total}} =
\mathcal{L}_{\text{img}}
+ \mathcal{L}_{\text{gain}}.
\end{equation}

The framework is trained in two phases. In Phase~1, the diffusion module with LoRA and the HPGM network are jointly trained to generate the base SDR image and the corresponding gain map. In Phase~2, the model is further optimized using MEF outputs generated by the Phase~1 model as the base SDR input, which reduces the domain gap between training and inference. This two-phase design allows HPGM to adapt from GT-anchored SDRs to MEF-generated SDRs encountered at inference time. The same objective function is used in both phases.
\section{Experiments}
\subsection{Datasets}
We conduct experiments on three publicly available multi-exposure benchmarks, including the Kalantari dataset\cite{kalantari2017deep} with 74 training and 15 test scenes, Tel dataset\cite{tel2023sctnet} with 108 training and 36 test scenes, and Challenge123\cite{kong2024safnet} with 288 training and 81 test scenes. These datasets contain both static and dynamic scenes with varying motion patterns and exposure differences. For a unified evaluation setting, we combine the official training splits of all three datasets for model training and merge their respective test splits for evaluation.

\subsection{Evaluation Metrics}
We evaluate performance using both full-reference and no-reference metrics. For distortion-based fidelity, we report PU-PSNR and PU-SSIM\cite{wang2004ssim} in the perceptually uniform domain. Perceptual similarity is further assessed using LPIPS\cite{zhang2018lpips} and DISTS\cite{ding2020dists}. In addition, we compute no-reference image quality metrics, including CLIPIQA\cite{wang2023clipiqa}, NIQE\cite{mittal2012niqe}, and BRISQUE\cite{mittal2012brisque}. For fair evaluation in HDR settings, no-reference metrics are computed after applying $\mu$-law compression.

\subsection{Implementation Details}

Our framework is implemented in PyTorch and trained on a single NVIDIA A6000 GPU. 
In Phase~1, the LoRA-adapted diffusion backbone and HPGM are jointly trained for 30k iterations to learn SDR synthesis and gain prediction. In Phase~2, HPGM is further optimized for 30k iterations using MEF outputs generated by the Phase~1 model. 
LoRA is applied to the cross-attention projections (\texttt{to\_q}, \texttt{to\_v}) with rank 8. 
Training uses random $512\times512$ crops with the Adam optimizer. The learning rates are $3\times10^{-5}$ for LoRA and $5\times10^{-5}$ for HPGM. 
For HDR supervision, $\mu$-law compression ($\mu=5000$) is used with loss weights $w_{\text{qmax}}=0.3$ and $\lambda_{\text{perc}}=10^{-2}$.

\begin{table}[t]
\centering
\caption{Quantitative comparison on the Challenge123, Tel, and Kalantari test datasets for HDR reconstruction methods. For fair comparison, NR-IQA metrics are computed on $\mu$-law tone-mapped HDR images. Best results are in bold and second-best results are underlined. 
($\uparrow$ indicates higher is better, $\downarrow$ indicates lower is better.)}
\label{tab:quant_results}
\resizebox{\linewidth}{!}{
\begin{tabular}{l|ccccccc}
\toprule
Method 
& PU-PSNR$\uparrow$ 
& PU-SSIM$\uparrow$ 
& LPIPS$\downarrow$ 
& DISTS$\downarrow$ 
& CLIPIQA$\uparrow$ 
& NIQE$\downarrow$ 
& BRISQUE$\downarrow$ \\
\midrule
AHDRNet        & 41.99 & 0.9780 & 0.0438 & 0.0309 & 0.3963 & 3.61 & 24.78 \\
HDR-GAN        & 42.74 & 0.9782 & 0.0457 & 0.0286 & 0.3957 & 3.61 & 25.23 \\
HDR-Transformer& 43.74 & 0.9806 & 0.0382 & 0.0232 & 0.3982 & 3.62 & 24.12 \\
SCTNet         & 42.92 & 0.9783 & 0.0399 & 0.0246 & 0.3959 & \textbf{3.56} & 23.84 \\
SAFNet         & 42.98 & 0.9802 & 0.0344 & 0.0213 & 0.4019 & 3.63 & \underline{23.12} \\
DiffHDR        & 42.50 & 0.9739 & 0.0464 & 0.0297 & \underline{0.4101} & 3.86 & 26.15 \\
RFG-HDR        & 43.16 & 0.9788 & 0.0367 & 0.0239 & 0.3927 & \underline{3.59} & 23.98 \\
AFUNet         & \underline{44.48} & \underline{0.9875} & \underline{0.0205} & \underline{0.0162} & 0.3983 & 3.68 & 27.13 \\

\textbf{Ours}  & \textbf{44.63} & \textbf{0.9878} & \textbf{0.0156} & \textbf{0.0147} & \textbf{0.4137} & \textbf{3.56} & \textbf{22.18} \\
\bottomrule
\end{tabular}
}
\end{table}

\begin{figure*}[t]
\centering
\includegraphics[width=\linewidth]{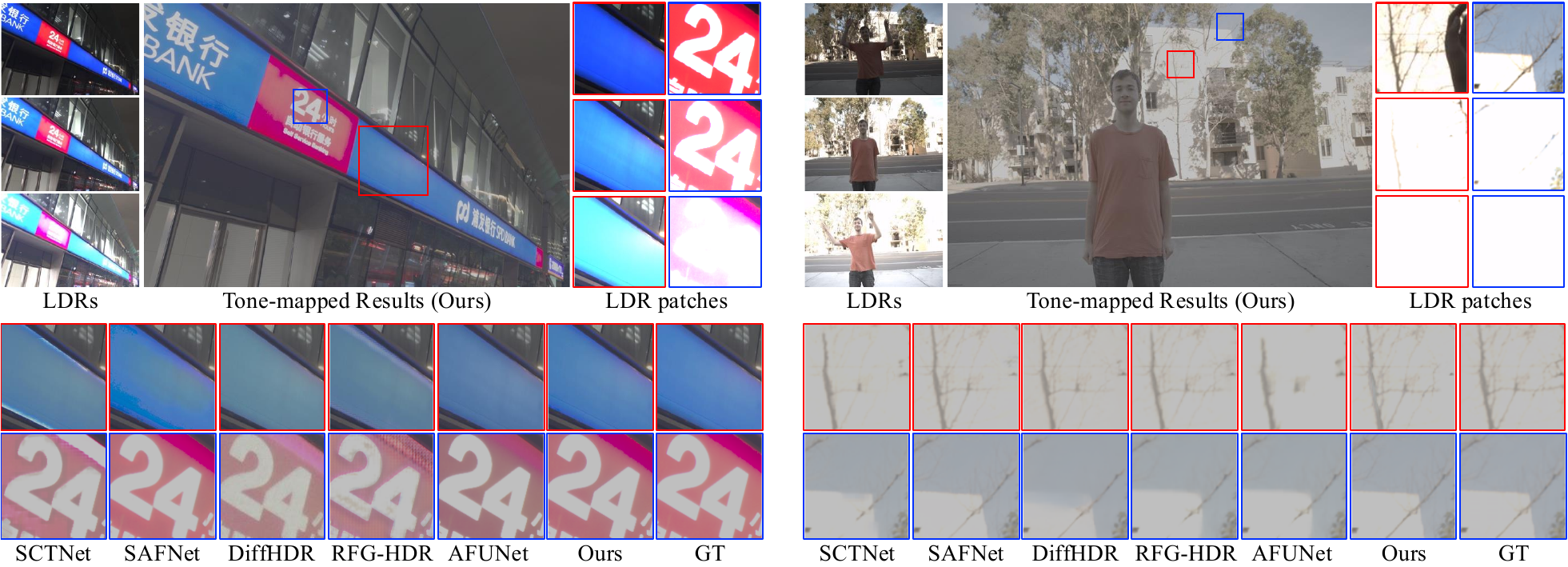}
\caption{
Qualitative comparison with HDR reconstruction methods. 
Our method produces visually pleasing PQ tone-mapped HDR results with improved highlight recovery and richer color appearance compared to existing approaches.
}
\label{fig:qual_hdr}
\end{figure*}

\subsection{Comparison on HDR Reconstruction}

Table~\ref{tab:quant_results} presents quantitative comparisons with HDR reconstruction methods on the Challenge123\cite{kong2024safnet}, Tel\cite{tel2023sctnet}, and Kalantari\cite{kalantari2017deep} test datasets. 
For fair evaluation, the no-reference scores are computed on $\mu$-law tone-mapped HDR images.
The proposed method achieves the best overall performance.
In particular, our approach achieves better reconstruction quality compared with previous methods. 
The results also show consistent improvements in perceptual quality across multiple evaluation measures. 
Overall, the proposed framework consistently outperforms previous methods across diverse evaluation metrics.

In addition, from a qualitative perspective, our method produces more visually consistent tone-mapped HDR results compared with existing methods as illustrated in Fig.~\ref{fig:qual_hdr}. In particular, the signboard regions exhibit more vivid color reproduction, while fine details in the sky and tree branches are better preserved.

\subsection{Comparison on Multi-Exposure Fusion}

\begin{figure}[t]
\centering

\begin{tabular*}{\linewidth}{@{\extracolsep{\fill}}cc}

\begin{minipage}{0.55\linewidth}
\centering
\includegraphics[width=\linewidth]{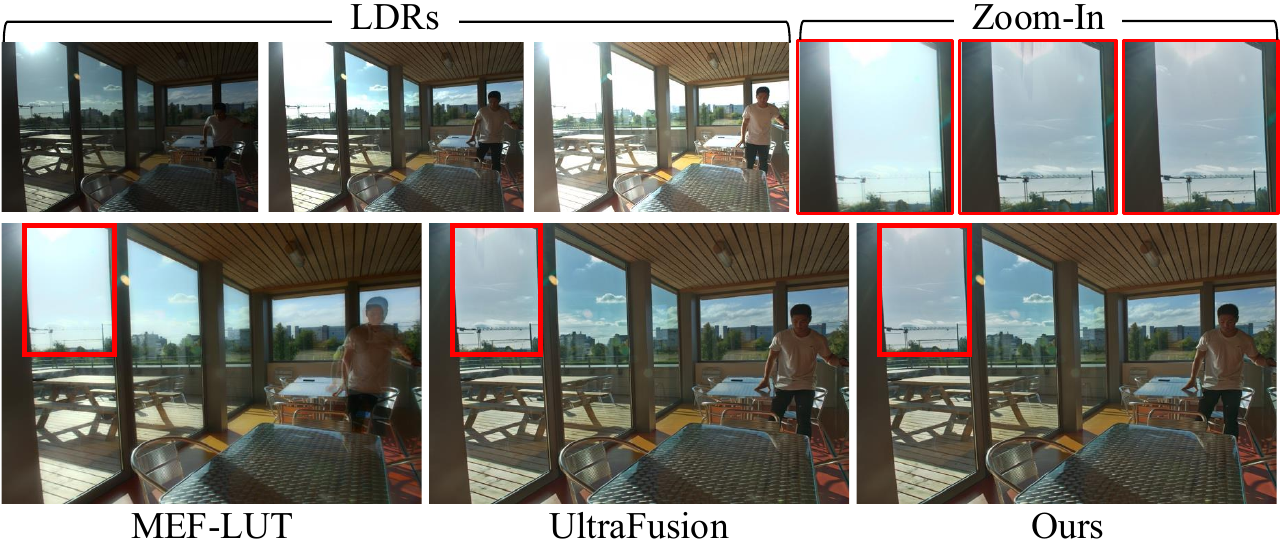}
\caption{Qualitative comparison with MEF methods. Our method produces more natural fusion with fewer highlight artifacts.}
\label{fig:qual_mef}
\end{minipage}
&
\begin{minipage}{0.42\linewidth}
\centering
\captionof{table}{
NR-IQA comparison on MEF methods over the Challenge123, Tel, and Kalantari datasets. All methods take three exposure images as input. Best results are shown in \textbf{bold}, and second-best results are \underline{underlined}.
}
\label{tab:noref_results}

\resizebox{\linewidth}{!}{
\begin{tabular}{lcccc}
\toprule
Method 
& MANIQA$\uparrow$ 
& CLIPIQA$\uparrow$ 
& NIQE$\downarrow$ 
& BRISQUE$\downarrow$ \\
\midrule
MEF-LUT        & 0.3650 & 0.4070 & 3.92 & 36.53 \\
UltraFusion    & 0.4064 & 0.5147 & \textbf{3.34} & 17.01 \\
\textbf{Ours}  & \textbf{0.4129} & \textbf{0.5233} & \underline{3.37} & \textbf{16.52} \\
\bottomrule
\end{tabular}
}
\end{minipage}

\end{tabular*}

\end{figure}

Table~\ref{tab:noref_results} and Fig.~\ref{fig:qual_mef} compare the proposed method with representative MEF approaches. 
For fair comparison, all methods use the same three exposure images as input to generate a fused SDR image.

As shown in Table~\ref{tab:noref_results}, the proposed method achieves competitive performance on most perceptual metrics, and is comparable to or outperforms existing MEF approaches. This demonstrates that the proposed framework preserves the strong fusion capability of the original UltraFusion backbone while benefiting from LoRA-based adaptation.
Qualitative comparisons in Fig.~\ref{fig:qual_mef} further show that the proposed method produces more visually pleasing fusion results with improved contrast, while reducing artifacts in bright sunlight regions.

Importantly, unlike conventional MEF approaches that only generate SDR outputs, our method additionally enables HDR reconstruction through the subsequent gain map estimation stage, providing a unified framework for both perceptual fusion and HDR reconstruction.

\subsection{Ablation Study}

\subsubsection{Comparison with a Cascaded Pipeline.}

As we previously mentioned in the introduction, one might view our reformulation of jointly predicting SDR and HDR as a simple concatenation of individual models. However, we observed that such a naïve cascaded pipeline is suboptimal in practice. To evaluate the effectiveness of the proposed unified framework, we compare it against a cascaded baseline that sequentially combines UltraFusion for SDR synthesis and GMNet\cite{liao2025learning} for gain map prediction. Specifically, the SDR output produced by pretrained UltraFusion\cite{chen2025ultrafusion} is directly fed into a pretrained GMNet, and both models are used as-is without any additional joint optimization.

As shown in Table~\ref{tab:pipeline_compare}, the cascaded pipeline yields substantially lower reconstruction quality. This result indicates that naively combining independently trained modules is insufficient for reliable HDR reconstruction. In contrast, our framework jointly learns SDR synthesis and gain map estimation within a unified pipeline, leading to significantly improved fidelity and perceptual consistency.

\subsubsection{Cross-Attention Comparison.}

Table~\ref{tab:ablation_crossattn} compares the proposed cross-attention design with the feature fusion strategy used in UltraFusion. UltraFusion combines guidance features from different exposures through normalized summation before the attention module. 
In contrast, our method processes exposure-specific guidance through separate cross-attention branches.

As shown in Table~\ref{tab:ablation_crossattn}, the proposed design consistently improves reconstruction quality across all metrics, achieving higher PU-PSNR and PU-SSIM as well as lower LPIPS and DISTS. 
These results suggest that preserving exposure-specific information during attention leads to more effective feature integration.

\begin{table*}[t]
\centering

\begin{minipage}[t]{0.49\linewidth}
\centering
\caption{Comparison with a pipeline combining UltraFusion and GMNet. Both models are pretrained.}
\label{tab:pipeline_compare}

\resizebox{\linewidth}{!}{
\begin{tabular}{lcccc}
\toprule
Method & PU-PSNR$\uparrow$ & PU-SSIM$\uparrow$ & LPIPS$\downarrow$ & DISTS$\downarrow$ \\
\midrule
UltraFusion+GMNet & 22.96 & 0.8688 & 0.1042 & 0.0765 \\
\textbf{Ours} & \textbf{44.63} & \textbf{0.9878} & \textbf{0.0156} & \textbf{0.0147} \\
\bottomrule
\end{tabular}
}
\end{minipage}
\hfill
\begin{minipage}[t]{0.49\linewidth}
\centering
\caption{Ablation study on cross-attention design with three exposure inputs.}
\label{tab:ablation_crossattn}

\resizebox{\linewidth}{!}{
\begin{tabular}{lcccc}
\toprule
Method & PU-PSNR$\uparrow$ & PU-SSIM$\uparrow$ & LPIPS$\downarrow$ & DISTS$\downarrow$ \\
\midrule
UltraFusion & 43.76 & 0.9876 & 0.0180 & 0.0157 \\
Ours     & \textbf{44.63} & \textbf{0.9878} & \textbf{0.0156} & \textbf{0.0147} \\
\bottomrule
\end{tabular}
}
\end{minipage}

\begin{minipage}[t]{0.37\linewidth}
\centering
\caption{Ablation study on input configurations during HDR prior construction.}
\label{tab:ablation_inputs}

\resizebox{\linewidth}{!}{
\begin{tabular}{lcccc}
\toprule
Inputs & PU-PSNR$\uparrow$ & PU-SSIM$\uparrow$ & LPIPS$\downarrow$ & DISTS$\downarrow$ \\
\midrule
SDR only & 23.69 & 0.8487 & 0.1310 & 0.0846 \\
UE, ME   & 25.32 & 0.8794 & 0.1126 & 0.0732 \\
OE, ME   & 26.16 & 0.8920 & 0.1041 & 0.0701 \\
All            & \textbf{44.63} & \textbf{0.9878} & \textbf{0.0156} & \textbf{0.0147} \\
\bottomrule
\end{tabular}
}
\end{minipage}
\hfill
\begin{minipage}[t]{0.60\linewidth}
\centering
\caption{
Ablation study on the two-phase training strategy.
}
\label{tab:two_phase_ablation}

\resizebox{\linewidth}{!}{
\begin{tabular}{l|ccccccc}
\toprule
Method 
& PU-PSNR$\uparrow$ 
& PU-SSIM$\uparrow$ 
& LPIPS$\downarrow$ 
& DISTS$\downarrow$ 
& CLIPIQA$\uparrow$ 
& NIQE$\downarrow$ 
& BRISQUE$\downarrow$ \\
\midrule
AFUNet 
& \underline{44.48} 
& \underline{0.9875} 
& 0.0205 
& 0.0162 
& 0.3983 
& 3.68 
& 27.13 \\
\midrule 
Phase 1 only
& 40.53 
& 0.9870
& \underline{0.0162} 
& 0.0161 
& \textbf{0.4180} 
& 3.62 
& \textbf{21.60} \\
Phase 2 only
& 43.26 
& 0.9864 
& 0.0178 
& \textbf{0.0143} 
& 0.4114 
& \underline{3.57} 
& 22.77 \\
\textbf{Phase 1+2} 
& \textbf{44.63} 
& \textbf{0.9878} 
& \textbf{0.0156}
& \underline{0.0147} 
& \underline{0.4137} 
& \textbf{3.56} 
& \underline{22.18} \\
\bottomrule
\end{tabular}
}
\end{minipage}

\end{table*}

\subsubsection{HPGM Input Comparison.}

We analyze the effect of different exposure inputs used for HDR prior construction in Table~\ref{tab:ablation_inputs}. 
Without exposure guidance, the model relies solely on the base SDR input, resulting in significantly degraded performance. 
When partial exposure information (UE, ME) or (OE, ME) is included, the reconstruction quality improves, indicating that complementary exposure cues help stabilize gain estimation.
However, using only two exposures still limits the available radiometric information.
Using all three exposures yields the best results across all metrics. 
This confirms that combining under-, mid-, and over-exposed images provides complementary information for reliable HDR prior construction and accurate gain map estimation.

\subsubsection{Training Phase Comparison.}

We evaluate the effectiveness of the proposed two-phase training strategy.
As shown in Table~\ref{tab:two_phase_ablation}, training only Phase~1 already produces competitive results, indicating that the end-to-end training of the SDR synthesis and gain estimation modules provides a strong initialization. 
Phase~2 alone also improves several perceptual metrics, particularly reducing DISTS, suggesting that it helps refine gain estimation for SDR images generated by the MEF process.
However, the best performance is achieved when both phases are combined. 
The full Phase~1+2 model consistently yields the best overall results across most metrics, even surpassing AFUNet\cite{li2025afunet}.
These results show that each phase of the proposed training strategy contributes to performance improvement. 

\subsection{Tone-mapped Images vs. Base SDR Results}

Conventional HDR reconstruction methods\cite{li2025afunet,kong2024safnet,tel2023sctnet} produce linear HDR images that require tone mapping for visualization on standard displays. 
Since the perceptual appearance of HDR images strongly depends on the selected tone mapping operator, the final visual quality may vary across different operators and display conditions. Moreover, as conventional tone mapping operators typically do not explicitly model scene content, they may yield suboptimal perceived visual quality.
In contrast, our framework directly synthesizes a perceptually balanced, display-ready SDR image during the MEF stage.

\begin{figure*}[t]
\centering
\includegraphics[width=\linewidth]{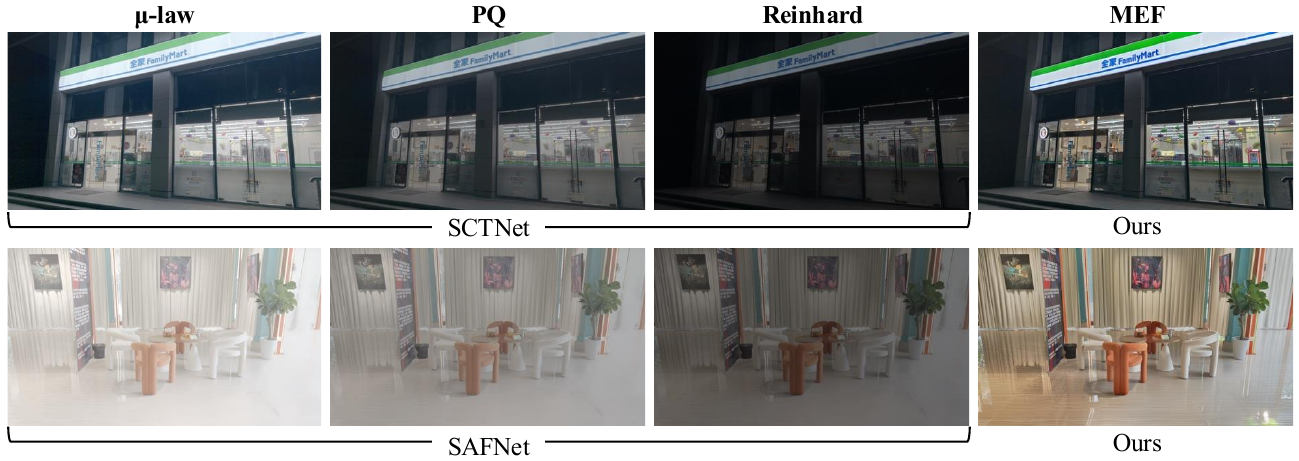}
\caption{
Comparison of different visualization operators ($\mu$-law, PQ, and Reinhard\cite{reinhard2023photographic}) and our SDR results. 
Our method directly produces a perceptually balanced SDR image, enabling consistent visualization without additional tone mapping.
}
\label{fig:tm_vs_mef}
\end{figure*}

To substantiate our claim, we compare the perceptual quality of tone-mapped HDR results with MEF-based SDR outputs. 
We apply several representative tone mapping operators to HDR results and compare them with the SDR images produced by our method.
As shown in \Cref{tab:tm_comparison} and \cref{fig:tm_vs_mef}, while the perceptual quality of existing HDR reconstruction methods varies depending on the selected tone mapping operator, our method produces visually pleasing results without requiring tone mapping. 
This observation supports that generating a high-quality base SDR representation leads to more stable and visually consistent results.

\begin{table}[t]
\centering
\caption{
NR-IQA comparison under different tone mapping operators ($\mu$-law, PQ, and Reinhard) for HDR reconstruction methods. Our method produces SDR outputs via multi-exposure fusion in Stage 1.}
\label{tab:tm_comparison}

\resizebox{\linewidth}{!}{
\begin{tabular}{l|ccc|ccc|ccc|c}
\toprule
& \multicolumn{3}{c|}{\bfseries AFUNet}
& \multicolumn{3}{c|}{\bfseries SCTNet}
& \multicolumn{3}{c|}{\bfseries SAFNet}
& \bfseries Ours \\
\cmidrule(lr){2-4} \cmidrule(lr){5-7} \cmidrule(lr){8-10} \cmidrule(lr){11-11}
\bfseries Metric
& $\mu$-law & PQ & Reinhard
& $\mu$-law & PQ & Reinhard
& $\mu$-law & PQ & Reinhard
& MEF \\
\midrule

CLIPIQA$\uparrow$
& 0.3894 & 0.3590 & 0.3708
& \underline{0.3984} & 0.3653 & 0.3719
& 0.3964 & 0.3585 & 0.3606
& \textbf{0.4984} \\

NIQE$\downarrow$
& 3.66 & 3.97 & 4.53
& 3.61 & 3.98 & 4.55
& \underline{3.55} & 3.89 & 4.48
& \textbf{3.53} \\

BRISQUE$\downarrow$
& 26.39 & 30.11 & 38.10
& \underline{22.65} & 26.13 & 35.82
& 23.36 & 27.59 & 36.59
& \textbf{19.88} \\

\bottomrule
\end{tabular}
}
\end{table}

\section{Conclusion}
We introduced DOME-HDR, a dual-output multi-exposure HDR framework that couples LoRA-adapted diffusion-based SDR fusion (with dual cross-attention) and HDR prior–guided gain map estimation (HPGM) in a single pipeline.
Unlike previous HDR imaging models that treat SDR fusion and gain map HDR reconstruction separately and assume a fixed SDR input, our unified framework jointly improves the SDR via diffusion-based fusion and estimates the gain map to produce both a pleasing SDR and its corresponding HDR.
Extensive evaluations demonstrate state-of-the-art HDR reconstruction performance, while retaining competitive SDR fusion quality.
\section*{Acknowledgements}
This research was supported by Artificial Intelligence Graduate School Program grant funded by Yonsei University (RS-2020-II201361), Samsung Research Funding \& Incubation Center of Samsung Electronics (SRFC-IT2501-02), Institute of Information \& Communications Technology Planning \& Evaluation (IITP) grant funded by the Korea government (MSIT): the Global AI Frontier Lab International Collaborative Research (RS-2024-00469482 \& RS-2024-00509279) and Development of Artificial Intelligence Technology for Self-Improving Competency-Aware Learning Capabilities (No.RS-2022-II220124).



%
%
\bibliographystyle{splncs04}
\bibliography{main}

\end{document}